\documentclass[MIRU,submit,english]{miru2025e}
\usepackage{graphicx}

\usepackage{stfloats}
\usepackage{tabularx} 
\usepackage[table,xcdraw,dvipsnames]{xcolor}
\usepackage{cite}
\usepackage{caption}

\usepackage[accsupp]{axessibility}
\usepackage[pagebackref,breaklinks,colorlinks]{hyperref}
\usepackage[capitalize]{cleveref}

\newcommand{\ie}{\textit{i}.\textit{e}.}
\newcommand{\eg}{\textit{e}.\textit{g}.}

\newcommand{\red}{\textcolor{red}}
\newcommand{\blue}{\textcolor{blue}}

\newcommand{\teal}{\textcolor{teal}}
\newcommand{\orange}{\textcolor{orange}}

\begin{document}

    \title{Selective Social-Interaction via Individual Importance for Fast Human Trajectory Prediction}

\affiliate{TTI}{Toyota Technological Institute}

 \author{Yota Urano}{TTI}[sd24413@toyota-ti.ac.jp]
 \author{Hiromu Taketsugu}{TTI}[sd25502@toyota-ti.ac.jp]
 \author{Norimichi Ukita}{TTI}[ukita@toyota-ti.ac.jp]

\maketitle

\section*{Abstract}
\label{abstract}
This paper presents an architecture for selecting important neighboring people to predict the primary person's trajectory.
To achieve effective neighboring people selection, we propose a people selection module called the Importance Estimator which outputs the importance of each neighboring person for predicting the primary person's future trajectory. 
To prevent gradients from being blocked by non-differentiable operations when sampling surrounding people based on their importance, we employ the Gumbel Softmax for training.
Experiments conducted on the JRDB dataset show that our method speeds up the process with competitive prediction accuracy.

\section{Introduction}
\label{intro}
Human trajectory prediction (HTP) is a crucial task for many applications. 
For instance, it is useful for autonomous systems such as autonomous vehicles~\cite{Trajectron++} and social robots~\cite{HST} to secure safe and efficient path planning~\cite{TP_survey}. 
In such kind of applications, HTP can be applied for collision avoidance with humans and autonomous robots.

\begin{figure}[t]
    \centering
    \includegraphics[width=1\linewidth]{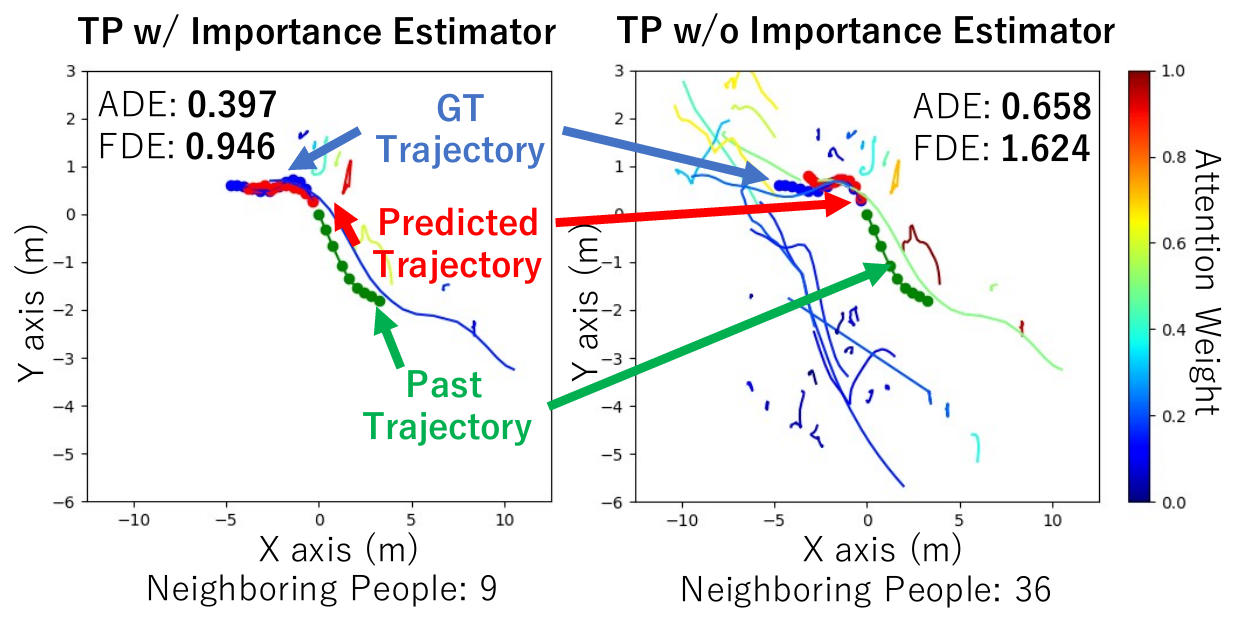}
    \caption{
    Trajectory prediction with and without our Importance Estimator: The left figure shows the prediction using only 9 neighboring people selected by our method, while the right figure uses all 36 people in the scene.
    The \teal{primary person’s past trajectory}, \red{predicted future trajectory}, and \blue{ground-truth trajectory} are shown in \teal{green}, \red{red}, and \blue{blue}, respectively.
    Thin lines represent the trajectories of surrounding people, with their
    attention weights.
    }
    \label{fig:championdata}
    \vspace{-10pt}
\end{figure}

One of the most important keys for predicting human trajectory is social-interaction~\cite{SocialForce,SocialLSTM,SocialGAN,Social-STGCNN,Eqmotion,Trajectron++,Social-Trans}.
However, especially when many people are involved in the scene, previous methods cannot handle social-interactions properly.
For example, as shown in Fig.~\ref{fig:championdata}, attention weights from the primary person to neighboring people are sometimes high even for people who are far from the primary person (right).
In contrast, we propose a method that can effectively ignore distant people (left), focusing only on important people relevant to the prediction.



To this end, we aim to correctly decrease the number of people for accurate prediction.
We introduce Importance Estimator,
a module that can select the people who are good or bad for predicting the primary person's trajectory.

During training, we employ Transformer masking to ignore attention to neighboring people, using this mechanism to train an importance estimator. At inference time, we apply a threshold to the estimator’s outputs to decide which people to select. Rather than masking out unimportant people, we simply omit them from the trajectory predictor’s inputs, eliminating unnecessary computations and reducing inference time. To address the non-differentiability of this selection process, we use a Gumbel–Softmax trick~\cite{GumbelSoftmax}. Additionally, to prevent all importance scores from collapsing to one, we introduce a loss term that maximizes the variance of the importance scores, thereby enabling fast and efficient inference.

Our contributions are threefold:
\begin{enumerate}
\item \textbf{Importance Estimator:}
We introduce an Importance Estimator that assigns each neighboring person an importance score of its influence on the primary person’s future trajectory. 
By selecting people based on their importance values, we can improve the processing speed while maintaining prediction accuracy.

\item \textbf{Gumbel–Softmax:}
We leverage the Gumbel–Softmax reparameterization~\cite{GumbelSoftmax} to sample discrete neighboring people selections during the forward pass while still propagating gradients through the underlying soft probability scores.

\item \textbf{Variance Loss:}
We augment the training objective with a variance-based loss term that discourages uniform importance scores. This encourages the model to differentiate truly influential people from less relevant ones, promoting diverse attention.
\end{enumerate}


\section{Related Work}
\label{related work}
\subsection{Social-interation for HTP.}
\label{social-interaction}
HTP has rapidly advanced in recent years, with many methods proposed to address it.~\cite{SocialForce,Flomo,FlowChain,Social-Trans,Emloco}.
Among them, approaches that consider interactions with neighboring people have been studied due to their significant impact on prediction performance. 

As an early model, Social Force model~\cite{SocialForce} attempts to avoid collisions by modeling social interaction as a repulsive force. 
Social-LSTM~\cite{SocialLSTM} enhances prediction accuracy by modeling interactions with neighboring people using social pooling.
EqMotion~\cite{Eqmotion} enables robust trajectory prediction by ensuring motion equivariance under Euclidean transformations and interaction invariance across neighboring people.
Various network architectures have been employed, including RNNs~\cite{SocialLSTM,Trajectron++}, Normalizing Flows~\cite{FlowChain,Flomo}, Diffusion models~\cite{MID,Singulartrajectory}, and Transformers~\cite{Social-Trans,Emloco}.
Among these, Transformers treat a variable number of people as tokens, and their attention mechanism enables the model to learn the relative importance of each person for prediction.

However, this variable-length nature introduces increased computational costs in modeling interactions.
To address this issue, this study aims to accelerate prediction while maintaining accuracy by selecting only a small number of relevant neighboring people for interaction modeling.

\begin{figure*}[!t]
    \centering
    \includegraphics[width=\linewidth]{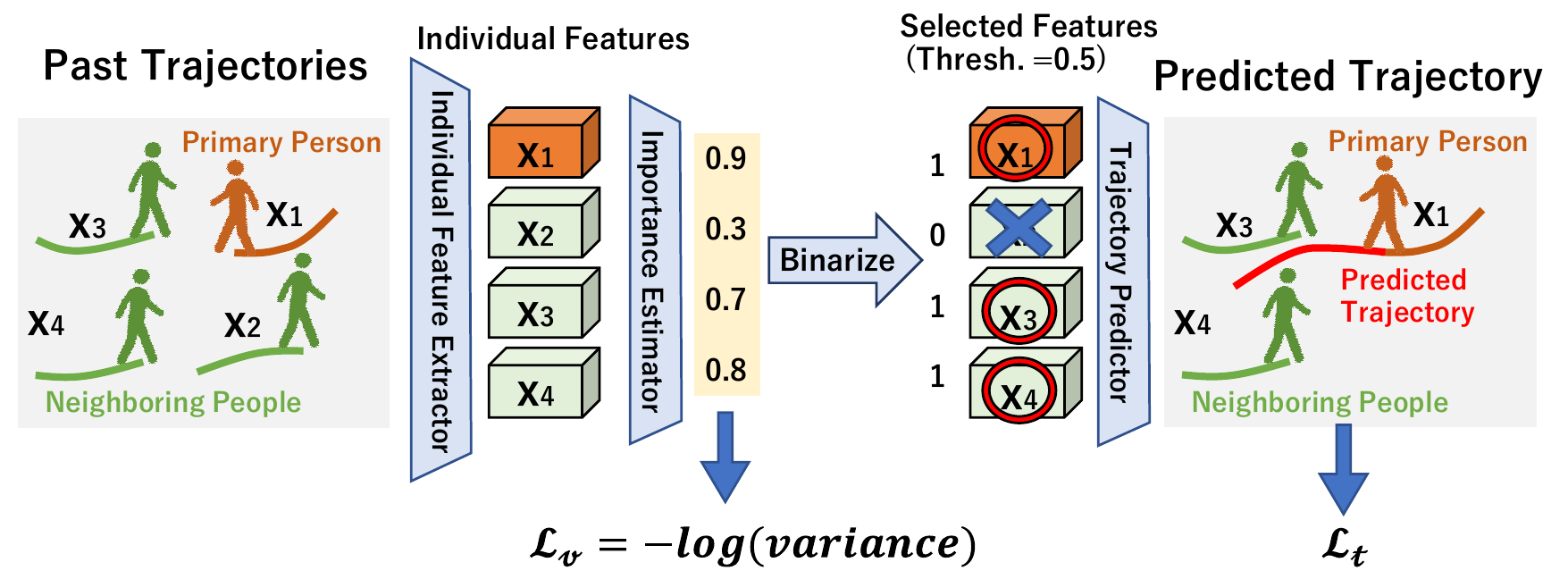}
    \caption{Overview of our proposal method. We extract individual features from past trajectories using the Individual Feature Extractor and the Importance Estimator computes importance scores for neighboring people. We leverage the Gumbel–Softmax reparameterization to enable differentiable sampling of discrete neighboring people selections. Binarized importance scores are utilized for thresholding to select only the small number of influential people at inference. Selected features are fed into the Trajectory Predictor to forecast future trajectories. During training, we also include a variance‐based loss term to encourage diversity in the importance scores and prevent collapse.}
    \label{fig:overview}
\end{figure*}

\subsection{Inference Acceleration for HTP}
\label{fast inference}
In applications such as autonomous driving and service robotics, the speed of HTP is crucial. 
To this end, various methods have been proposed to accelerate prediction~\cite{FlowChain,Social-STGCNN}. 

FlowChain~\cite{FlowChain} employs normalizing flows to enable fast predictions of each person's future positions. Furthermore, it improves inference speed by shared computations across people.
Social-STGCNN~\cite{Social-STGCNN} achieves fast prediction by simultaneously extracting trajectory features and modeling person interactions through graph convolutional networks.

However, these methods typically input all people in the scene into the model and predict future trajectories based on all pairwise interactions. This comprehensive modeling can lead to slower inference.
To overcome this, we propose a lightweight importance estimator that selects only relevant people for prediction, enabling the trajectory predictor to forecast more efficiently with fewer neighboring people.

\section{Method}
\label{method}
We propose a model that outputs an importance score for each neighboring person, indicating how relevant they are to predicting the trajectory of a primary person. 
In this section, we explain a problem formulation (Sec.~\ref{problem formulation}), Importance Estimator (Sec.~\ref{importance estimator}), how to learn an Importance Estimator (Sec.~\ref{learning method}), and a variance loss function (Sec.~\ref{loss function}).

Figure~\ref{fig:overview} shows the overview of the proposed method. By selecting neighboring people whose scores exceed a threshold, the trade-off between prediction accuracy and inference speed can be adjusted to suit different use cases.

\subsection{Problem Formulation}
\label{problem formulation}
We represent the trajectory sequence of person $i$ by $X_i$ and $Y_i$ which mean past and future trajectory, respectively. 
The observed time-steps are defined as $t = 1, ..., T_{obs}$, while the prediction time-steps are defined as $t = T_{obs}+1, ..., T_{pred}$.
In a scene with $N$ people, the network input is $X$ = [$X_1$, $X_2$, $X_3$, ... , $X_N$], output is $Y$ = [$Y_1$, $Y_2$, $Y_3$, ..., $Y_N$], and ground-truth is $\hat{Y}$ = [$\hat{Y_1}$, $\hat{Y_2}$, $\hat{Y_3}$, ..., $\hat{Y_N}$]. The primary person's index is always 1.

\subsection{Importance Estimator}
\label{importance estimator}
The importance estimator calculates how much each of the observed neighboring people $i$ contributes to the primary person’s future trajectory $Y_1$. Concretely, it takes each person’s features -such as past trajectories $X_i$ and relative positions— as input and uses mechanisms like attention to compute a scalar ``importance score''. Based on these scores, we can decide the threshold to limit the number of neighboring people fed into the predictor: by adjusting the score threshold, one can choose to prioritize computational speed or predictive accuracy. This approach reduces the input dimensionality (\ie, the number of people considered) while retaining only the most influential interactions.

\subsection{Learning Importance Estimator}
\label{learning method}
The selection operation is non-differentiable because it involves discrete decisions \eg, choosing whether to retain or ignore a feature.
As a result, gradients cannot flow through, thus hindering the training of the importance estimator.
To address this issue, we adopt a gradient propagation technique using Gumbel Softmax~\cite{GumbelSoftmax}.
During the forward pass, sampling is performed according to the probability distribution using the Gumbel Softmax.
Then, in the backward pass, instead of propagating gradients through the sampled discrete scores, gradients are passed through the soft probability scores.
This trick enables training by allowing gradients from HTP loss to flow through the sampling process.
After the Trajectory Predictor is trained by a trajectory loss function, the Importance Estimator is  trained to avoid interfering with the training of the Trajectory Predictor.
The trajectory loss function $\mathcal{L}_t$ is defined as the mean squared error (MSE) between the predicted and ground-truth future trajectory. 
The loss is formulated as follows:
\begin{equation}
    \mathcal{L}_t = \frac{1}{T_{pred}-T_{obs}-1} \sum_{t=T_{obs}+1}^{T_{pred}} \| \hat{\mathbf{Y}}_i^t - \mathbf{Y}_i^t \|^2
\end{equation}

\subsection{Variance Loss Function}
\label{loss function}
Inspired by the training strategy of DINO~\cite{DINO}, we design a loss function that prevents the collapse of importance scores to a single constant value, thereby promoting diversity. Specifically, we introduce a loss term that decreases as the variance of the importance scores increases, which helps avoid a non-discriminative state where all people are treated equally. This encourages the model to focus more on truly influential people.
The loss is formulated as follows:
\begin{equation}
    \mathcal{L}_v = -log(variance+\epsilon)
\end{equation}
where $\epsilon$ is a small constant to prevent the loss from exploding when the variance becomes zero. 

As a result, this loss is added to the trajectory loss $\mathcal{L}_t$ with a weighting factor $\alpha$ as follows:
\begin{equation}
    \mathcal{L}_{total} = \mathcal{L}_t + \alpha \mathcal{L}_v
\end{equation}
where $\alpha=1$ serves to balance the two loss terms.

\section{Experiments}
\label{experiments}
In this section, we present the experimental setups and analyze the effectiveness of our method.

\begin{figure}[t]
    \centering
    \includegraphics[width=1\linewidth]{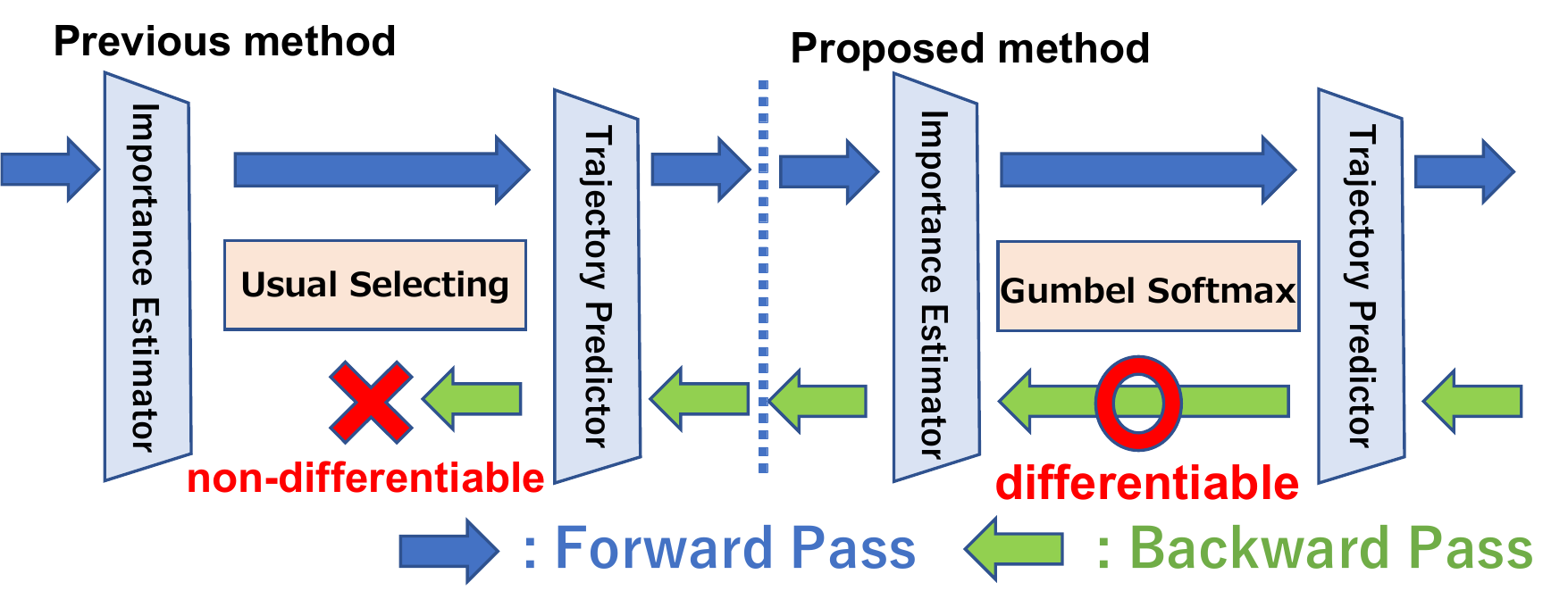}
    \caption{Gumbel Softmax :
    Usual selection operations are non-differentiable, preventing gradients from flowing from the Trajectory Predictor back to the Importance Estimator.
    In contrast, by using Gumbel Softmax, we can perform discrete sampling during the forward pass, while still allowing gradients to flow through the original probability scores during the backward pass, making the process differentiable.}
    \vspace{-5mm}
    \label{fig:enter-label}
\end{figure}

\subsection{Experimental Setup}
\label{experimental}
\textbf{Dataset.}
\label{dataset}
We use JackRabbot Dataset and Benchmark (JRDB)~\cite{JRDB}. 
JRDB is a real-world dataset that provides a wide variety of person trajectories and 2D bounding boxes.
We predict the locations for the next 12 time steps given the past 9 time steps at 2.5 fps.
We use a subset provided by the Social-Transmotion (Social-Trans.)~\cite{Social-Trans}.
Because JRDB has a small number of pose information, we only use people's trajectories and 2D bounding boxes.

\textbf{Metrics and a baseline.}
\label{metrics}
We evaluate the models in terms of Average Displacement Error (ADE), Final Displacement Error (FDE), and FLoating-point OPerations (FLOPs).
ADE/FDE is an average/final displacement error between the predicted and ground-truth locations of the person. 
FLOPs count the total number of floating-point arithmetic operations (\eg, multiplications and additions) executed by a model during inference, providing a hardware-agnostic measure of its computational complexity.

As a baseline, we adopt Social-Trans~\cite{Social-Trans}, a Transformer-based model that captures spatiotemporal social 
-interactions by representing each person’s trajectory as a sequence of tokens. Its self-attention mechanism enables dynamic modeling of inter-person interactions without requiring any map or visual inputs.

\textbf{Implementation details.}
\label{Impli}
The Importance Estimator’s architecture comprises three linear layers and a compact Transformer. First, linear layers map each person’s feature vector —produced by the Social-Trans~\cite{Social-Trans} individual feature extractor— to a 64-dimensional embedding. A small Transformer then computes how much attention the primary person allocates to each neighboring person, yielding raw importance signals. Finally, the Transformer outputs are passed through a linear layer to produce a score for each neighboring person, and a sigmoid activation maps these scores to [0, 1]. During inference time, we use 0.5 as a threshold of an importance value.

\subsection{Quantitative evaluation}
\label{Quantitative}
\begin{figure}[!t]
    \centering
    \includegraphics[width=1\linewidth]{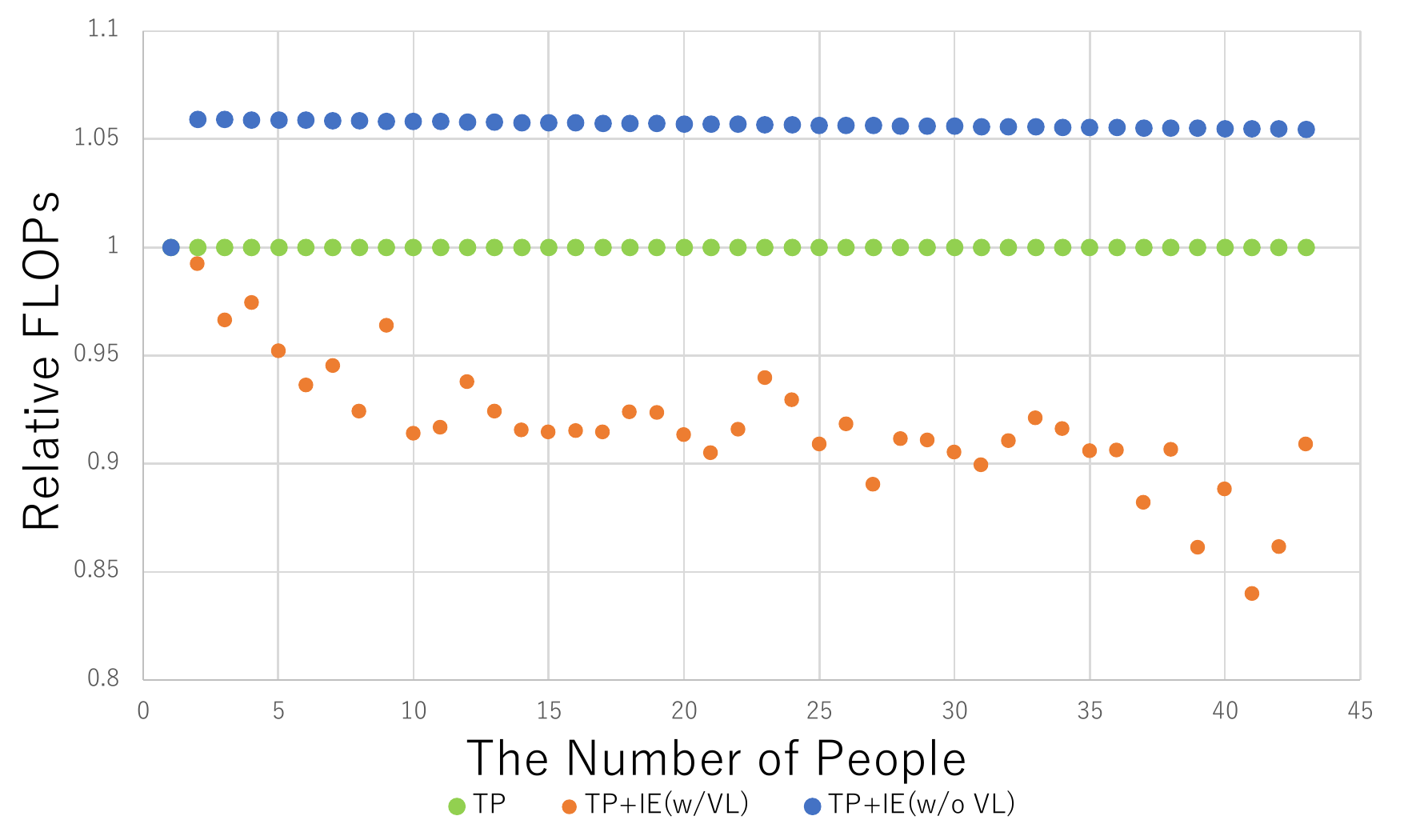}
    \caption{Comparison of FLOPs by number of people in the scene, with \teal{the Trajectory Predictor (TP) alone} is shown in \teal{green} and \orange{TP combined with the Importance Estimator (IE)} is shown in \orange{orange}. As an ablation study, the FLOPs of \blue{TP and IE without the variance loss} are shown in \blue{blue}. Each FLOPs value is computed relative to the trajectory prediction, which served as the denominator.}
    \label{fig:FLOPs}
\end{figure}

\textbf{Inference speed.}
\label{InferenceSpeed}
To demonstrate the impact of the Importance Estimator on the trajectory predictor, we compare the Transformer-based predictor with and without the estimator attached.
Because the FLOPs of the Transformer scale with input length, we conduct this comparison for each input dimension. 
The results are shown in Figure~\ref{fig:FLOPs}. Relative FLOPs for the trajectory predictor alone are plotted in green, while FLOPs for the predictor augmented with the Importance Estimator are plotted in orange. FLOPs for each setting are calculated as a ratio, using the trajectory prediction result as the denominator.
All measurements were taken with a batch size of 1. 
Compared to the green line, the gap between the orange and green curves widens with larger person counts—reflecting greater savings when more people can be pruned. 
Conversely, when the scene contains few people, only a small number can be removed, and the overhead introduced by the Importance Estimator causes both configurations to exhibit similar FLOPs.
As an ablation study, we also present results obtained by training the Importance Estimator using only the trajectory-prediction loss, omitting the variance loss (``VL''). 
Relative to the vanilla TP, using IE without VL leads to a 5.7\% (1.49G FLOPs to 1.59G FLOPs) average increase in FLOPs, whereas using IE with VL results in an 8.1\% (1.49G FLOPs to 1.37G FLOPs) average reduction.
In this configuration, the FLOPs increase by exactly the estimator’s overhead, indicating that all importance scores collapse to one and that every person in the scene is retained for prediction.

\textbf{Prediction accuracy.}
\label{PredictionAccuracy}
Table~\ref{fig:Accuracy} compares the prediction errors of the baseline Trajectory Predictor (``TP'')~\cite{Social-Trans} and the TP + Importance Estimator (``IE''). Integrating the IE increases ADE only from 0.376 to 0.377 ($ \approx $ 0.3\% relative) and FDE from 0.741 to 0.747 ($ \approx $ 0.8\% relative), despite using only influential people. These negligible degradations confirm that our method effectively reduces computational cost while preserving predictive accuracy.
The results demonstrate a favorable efficiency–performance trade-off, validating the practical utility of our approach.



\begin{table}[!t]
\centering
\caption{
Comparison of prediction errors between the baseline Trajectory Predictor (TP) and the TP + IE model, which reduces the number of input people. Despite using fewer inputs, both average displacement error (ADE) and final displacement error (FDE) remain nearly unchanged.
}
\label{fig:Accuracy}
\scalebox{1.2}{
\begin{tabular}{l|ll}
    & TP    & TP+IE    \\ \hline
ADE & 0.376 & 0.377 \\ \hline
FDE & 0.741 & 0.747
\end{tabular}
}
\end{table}


\section{Discussion}
\label{discussion}


\begin{table}[!t]
\centering
\caption{
Comparison between standard prediction (``Baseline'') and optimal exclusion strategy (``Oracle'').  
``Baseline'' uses all people in the scene for prediction, while ``Oracle'' selects the best prediction by masking each person one by one and choosing the most accurate outcome per scene. Although this Oracle setting is not feasible in real-world applications, it illustrates the theoretical upper bound of prediction accuracy improvement achievable by optimally selecting and excluding the most impactful person.
}
\vspace{2mm}
\label{tab:baseline_vs_oracle}
\scalebox{1.2}{
\begin{tabular}{l|ll}
    & Baseline & \multicolumn{1}{c}{Oracle} \\ \hline
ADE & 0.376   & 0.314                     \\ \hline
FDE & 0.741  & 0.589                    
\end{tabular}
}

\end{table}
To evaluate the potential benefit of excluding specific surrounding people, we compare two settings:  
``Baseline'', which denotes the standard prediction using all people in the scene, and  
``Oracle'', which denotes the best-case outcome obtained by masking each person one at a time and selecting the prediction with the highest accuracy for each test scene.  
As shown in Table~\ref{tab:baseline_vs_oracle}, Oracle outperforms Baseline, suggesting that selectively excluding them can lead to improved accuracy.

\section{Conclusion and Future Work}
\label{conclusion}
In this paper, we introduce the Importance Estimator, calculating the importance score of neighboring people to predict the primary person's future trajectory correctly.
 During training, we employ the Gumbel–Softmax reparameterization trick and augment the objective with a variance-based term to prevent those scores from collapsing to a single constant. 
Although our method succeeds in reducing inference time, it yields accuracy improvements only in certain cases. We attribute this limitation to an inherent conflict between maximizing score variance and enhancing prediction performance. To address this, we are exploring two remedies: (1) introducing an auxiliary loss term based on the ratio of prediction accuracies with and without neighboring people masking, and (2) incorporating a distance-based prior—i.e., weighting importance scores according to each person’s distance from the primary person. The detailed implementation of these extensions will be pursued in future work.



\clearpage
\bibliographystyle{miru2025e}
\bibliography{main}


\end{document}